\documentclass[aps,prd,twocolumn,nofootinbib]{revtex4-2}

\usepackage{graphicx}
\usepackage{amsmath,amssymb}
\usepackage{algorithm}
\usepackage{algpseudocode}
\usepackage{framed,multirow}
\usepackage{enumitem}
\usepackage{hyperref}
\usepackage{footnote}  

\begin{document}

\title{Commutative Evolution Laws in Holographic Cellular Automata:\\
AdS/CFT, Near-Extremal D3-Branes, and a Deep Learning Approach}

\author{Hyunju Go}
\affiliation{Chung-Ang University, Republic of Korea}
\email{h9555.go@gmail.com}

\date{December 31, 2024}

\begin{abstract}
According to ’t Hooft, restoring Poincaré invariance in a holographic cellular automaton (CA) requires two distinct evolution laws that commute.
We explore how this is realized in the AdS/CFT framework, assuming commutativity as a fundamental principle—much like general covariance once did—for encoding curvature.
In our setup, physical processes in a given spacetime are encoded in a CA; to preserve Poincaré symmetry, the spacetime curvature must effectively vanish, so we consider a near-extremal black D3-brane solution, in which both the stretched horizon and the conformal boundary are approximated by Minkowski space.
AdS/CFT implies a spatial evolution law connecting these hypersurfaces. 
Commutativity means the final state does not depend on the order of time evolution on each hypersurface and spatial evolution between them, forcing the time evolution law on the horizon and boundary to coincide.
To satisfy all these conditions, we aim to demonstrate that the spatial evolution law inevitably encapsulates the curvature of the bulk, including quantum effects.
For a computational model, we compactify the hyperplanes to tori, reducing the degrees of freedom to a finite number; taking these tori to infinite size then restores Poincaré symmetry.
We propose a deep learning algorithm that, given a known time evolution law and commutativity, deduces the corresponding spatial evolution law.
\end{abstract}

\maketitle

\section{Introduction}
The holographic principle is regarded as a fundamental aspect of quantum gravity theory.
This principle started from the fact that the entropy of a black hole is proportional to the area of the horizon and was extended to the general case\cite{Hooft_1}. 
Considering unitarity, entropy and counting arguments, it was pointed out that the 2-dimensional surface should determine all information in the 3+1 dimension. 
Note that considering the limit where the two-dimensional space becomes flat, this means that there is a transform in the direction perpendicular to that flat space.
Furthermore, since the ultimate theory will satisfy Poincaré invariance, this transformation and the time evolution will commute with each other. 
The proposed cellular automaton(CA) model, in which information in a two-dimensional space determines all information in the 3+1 dimension, had only a linear structure and could not include interactions.
Therefore, we aim to build a non-linear model from an AdS/CFT perspective.

The flat space-time used to construct this model was implemented as a holographic screen, and its connection with string theory was revealed\cite{Susskind}. 
An event horizon can in fact be mapped by following light rays.
Here, the focusing theorem guarantees that there exists at least one map satisfying the holographic principle.
This mapping unfolds the entropy on the black hole's horizon onto a flat surface, allowing black hole physics to remain on this screen.
However, because it does not evolve under Lorentzian time on a holographic screen, defining a CA model on this screen becomes difficult\cite{Bou}. 
Moreover, since the degree of freedom is not generally preserved, it is difficult to expect it to follow conventional physics.
Finally, in the AdS/CFT correspondence, the conformal boundary is realized as a holographic screen\cite{Malda_1}\cite{Witten_1}. 
Also, this correspondence guarantees the existence of a transformation that connects the event horizon and the conformal boundary.

We aim to take a step further by combining this remarkable fact with the commutative property and the CA model.
Specifically, considering the AdS/CFT correspondence, the main objective will be to explore how the argument used in 't Hooft's original paper\cite{Hooft_1} can be justified and to interpret the significance of the commutative property.
The motivation for focusing on commutativity in this paper is derived from 't Hooft's CA paradigm and is not identical to the standard renormalization group (RG) interpretation.
When two operations commute, their final outcome is independent of the order in which they are applied, thereby preventing path-dependent loss or distortion of information. 
In quantum mechanics, the uniqueness of mapping one initial state to a single final state precisely characterizes unitarity (i.e., information conservation). Hence, a conceptual link emerges such that “commutativity implies information preservation (unitarity).” Therefore, positing commutativity can be regarded as a plausible assumption necessary for maintaining information conservation.
However, this should not be taken to equate strictly to quantum-mechanical unitarity. 
True unitarity entails a norm-preserving linear operator on a Hilbert space. 
Our commutativity assumption is instead a toy-model principle for preventing path-dependent loss of information, rather than a complete statement of quantum unitarity.
Since defining the commutation relation on the boundary of a given spacetime using a CA model requires the boundary surfaces to be flat, we consider a near-extremal black D3-brane solution.
Accordingly, we will propose a possible interpretation about the commutation relations in AdS space, considering the thermal equilibrium state of a near-extremal black D3-brane.
One of the virtues of this holographic model is that it allows us to disregard the curvature within the given bulk.
Paradoxically, we will find that the transformation defined in the holographic direction within this interpretation rather encodes information about that curvature.

In the context of the black hole paradox, the piece of information inscribed on the horizon can be considered to be the encrypted information of the conformal boundary\cite{Malda}.
From this perspective, the function that goes from the horizon to the conformal boundary will be called the decrypting function.
Here, saying that this function is reversible is equivalent to saying that no information is lost in a black hole.
In this paper, we propose a computational model that can find this decrypting function.

Meanwhile, in the field of deep learning, there are three notable research results related to the general relativity, quantum field theory and AdS/CFT. 
Firstly, there has been a study on convolutional neural networks(CNN) in non-Euclidean space to effectively process data in non-Euclidean space such as social networks, medical information, brain imaging and computer graphics. 
For example, training a conventional CNN after projecting the data from the sphere onto a plane fails, 
but by constructing CNN that is invariant under rotation, it can be trained successfully with this data\cite{Group}\cite{Spherical}. 
Furthermore, this concept can be extended to include gauge symmetry and implemented through Icosahedral CNN\cite{Gauge}. 
As the concept of symmetry was introduced into CNN, elements necessary for convolution were mapped to geometric objects, 
and consequently, convolution could be expressed in covariant form\cite{General}\cite{Covariance}.

Also, there is a case of studying the AdS/CFT correspondance by mapping it to deep learning\cite{Koji_1}\cite{Koji_2}. 
In this study, the bulk metric function on the black hole horizon can be trained given the data at the boundary of the quantum field theory. 
The philosophy is that the bulk spacetime can correspond to a deep neural network.

Finally, since gauge/gravity theory and deep learning can be expressed as a matrix model, 
the correspondence between them have been considered\cite{Alexander:2021rch}. 
In this work, for restricted Boltzmann machine and cubic learning system, matter fields are mapped to a layer of given deep learning architecture and gauge/gravity fields are mapped to weight matrix. 
Furthermore, the learning process, which is the time-evolution of weight matrix, can be thought of as a process in which the effective law changes, such as spontaneous symmetry breaking.

These results imply that deep learning can be an effective computational tool even in quantum gravity. 
Therefore, the proposed computational model adopts a deep learning algorithm.

\section{Commutativity and AdS/CFT correspondence}
Here, we focus on the AdS black brane, thus bringing a many-body problem into consideration.
In this case, it is known that gauge theory above the conformal boundary has high temperature behavior\cite{Malda_1}\cite{Horo}\cite{Witten}.
Also, the corresponding black brane horizon can reach thermal equilibrium.
Therefore, any piece of information on a given boundary can be regarded as a microstate that evolves according to the time-evolution law while preserving degrees of freedom.
Additionally, for the construction of our computational AdS/CFT model, we assume that the microstates forming an ensemble are encoded in a cellular automaton, thus following the logic of the cellular automaton interpretation, 
but we will exclude fundamental discussions of quantum mechanics such as the Copenhagen interpretation or any other interpretational aspects of quantum mechanics\cite{Hooft_2}.
The holographic principle implies the existence of an evolution law in the holographic direction. 
This law will be called the $Z$ operator. 
Assuming that the invariance under the Poincaré group is recovered in the ultimate holographic theory, a certain time evolution law and $Z$ must commute\cite{Hooft_1}.
We will compare the information defined on the spatial slices of two spacetimes connected by $Z$.

To encode information about physical processes in a cellular automaton (CA) and maintain Poincaré symmetry, the spacetime under consideration must be Minkowski. Otherwise, it becomes difficult to define a time evolution law by slicing the spacetime into equal-time surfaces. Moreover, Poincaré symmetry itself is originally defined in Minkowski space.
In the context of AdS/CFT, two Minkowski hypersurfaces can be found in the near-extremal D3-brane solution in decoupling limit.
Maldacena’s near-extremal D3-brane metric (often referred to as Eq.~(2.5) in \cite{Malda_1}) is:
\begin{widetext}
\begin{equation}\tag{2.5}
ds^2
=
\sqrt{\frac{U^2}{4\pi gN}}
\Bigl[
 -\Bigl(1 - \tfrac{U_0^4}{U^4}\Bigr)\,dt^2
 + \sum_{i=1}^{3} dy_i\,dy^i
\Bigr]
\;+\;
\sqrt{\frac{4\pi gN}{U^2}}
\Bigl(1 - \tfrac{U_0^4}{U^4}\Bigr)^{-1} dU^2
\;+\;
\sqrt{4\pi gN}\,\mathrm{d}\Omega_5,
\end{equation}
\end{widetext}
where \(g\) is the string coupling, \(N\) is the number of D3-branes, and \(U_0\) sets the horizon scale.  
We assume \(gN\gg 1\) so that the supergravity approximation remains valid.
To see how the metric becomes approximately Minkowski near a \emph{stretched horizon}, consider fixing the radial coordinate at
\[
U = U_0 + \epsilon,
\quad
\epsilon \ll U_0,
\quad
\mathrm{d}U = 0 
\quad (\text{on this slice}).
\]
Since there is no radial variation on this slice, the term involving \(dU^2\) \emph{does not contribute} to the induced metric. Separately, we expand 
\[
1 - \frac{U_0^4}{U^4}
\;=\;
1 - \frac{1}{\Bigl(1 + \tfrac{\epsilon}{U_0}\Bigr)^4}
\;\approx\;
4\,\frac{\epsilon}{U_0}
\quad (\epsilon \ll U_0),
\]
and substitute into (2.5). At leading order in \(\epsilon\), the \((t,y_i)\) part simplifies to
\begin{equation}\label{eq:StretchedHorizonApprox}
ds^2 
\;\approx\;
\sqrt{\frac{(U_0+\epsilon)^2}{4\pi gN}}
\Bigl[
 -\,4\,\tfrac{\epsilon}{U_0}\,dt^2
 + \sum_{i=1}^{3} dy_i\,dy^i
\Bigr]
\;+\;
\sqrt{4\pi gN}\,\mathrm{d}\Omega_5,
\end{equation}
revealing a Minkowski-like form in the \((t,y_i)\) directions, up to an overall scale factor.
Meanwhile, in the limit \(U \to \infty\), we have 
\[
1 - \frac{U_0^4}{U^4}
\;\longrightarrow\;
1,
\]
and the metric becomes
\[
\begin{aligned}
ds^2
&~\approx~
\sqrt{\tfrac{U^2}{4\pi gN}}
\Bigl[
 -\,dt^2 
 + \sum_{i=1}^{3} dy_i\,dy^i
\Bigr]
\\
&\quad
+\,\sqrt{\tfrac{4\pi gN}{U^2}}\,dU^2
~+~
\sqrt{4\pi gN}\,\mathrm{d}\Omega_5\,,
\end{aligned}
\]
which becomes conformally equivalent to flat Minkowski space in the \((t,y_i)\) directions.
It is worth emphasizing that the approximate Minkowski form at the stretched horizon and at the asymptotic boundary applies only within these respective local patches. 
The bulk itself still possesses nontrivial curvature, and these Minkowski regions do not imply flatness throughout the entire spacetime. 
The time delay (or gravitational redshift) we will discuss arises because information must traverse a curved bulk region, despite the local near-flatness at the two boundaries.

\bigskip
\noindent
\textbf{Remark on Microstate Dynamics.}
Even though \(U_0\) is treated as a constant for this near-horizon analysis (i.e., a stationary black brane),  
the underlying \emph{microstates} associated with the brane can still evolve in time.  
In a holographic or cellular automaton (CA) interpretation, one may encode these microstates into configurations that change while the macro-scale parameter \(U_0\) remains fixed overall, maintaining thermal equilibrium.  
Hence, the bulk geometry at the stretched horizon can be nearly Minkowskian, yet the microscopic degrees of freedom can undergo nontrivial dynamics (e.g., Hawking-like processes balanced by an external bath), preserving the equilibrium at the macroscopic level.

On the other hand, when considering these two spacetimes together, we define the time variable merely as an index for a collection of space-like surfaces. 
Since the curvature of spacetime does not depend on the choice of coordinates, both spacetimes remain flat.
To delve deeper, consider a particular physical process at the stretched horizon. 
Information on the stretched horizon corresponding to that process can be transferred to the conformal boundary through Hawking radiation. 
Maybe, if certain particle falls into the horizon, this information can also be transferred by particles produced through quantum fluctuations, reaching the conformal boundary. 
Regardless of the mechanism, an observer at the conformal boundary will experience time dilation due to gravity.
However, for commutativity to hold in our assumed model, the law of time evolution must remain identical in both spacetime frames.
Let us denote this time evolution law by $T$.
Therefore, to satisfy these conditions, the time delay after a spatial translation $Z$ should not be expressed in the form of $T^n$, but rather, it should already be incorporated into $Z$.  
Although it may be hard to imagine, the CA model can produce a vast number of possible evolutions based on each piece of information, so it is reasonable to expect that there exist pairs of $Z$ and $T$ satisfying these conditions.

Let us return to the issue of information on the stretched horizon being transmitted to the conformal boundary. 
Note that a length scale is not defined on the conformal boundary.
By first matching the areas of the stretched horizon and the conformal boundary, we establish a common length scale for both planes. 
Next, consider a pair of short-wavelength photons produced in the bulk; one travels to the horizon, and the other to the boundary. 
Because each photon’s wavelength is sufficiently short, it can arrive at specific locations while carrying multiple bits of information—thus making the impact of the bulk’s curvature more transparent. 
In other words, the curved paths they follow lead them to distinct, asymmetrical arrival points, vividly illustrating how the bulk is distorted. 
In this scheme, the evolution law $Z$—which transfers information from the CA at the stretched horizon to the boundary—encodes the resulting displacement as a permutation of states. 
As a result, the difference in each photon’s endpoint, driven by the bulk curvature, is captured within $Z$ in the form of a rearrangement, effectively mapping the curvature into a permutation structure.  

However, as highlighted in discussions of entanglement entropy, the AdS/CFT correspondence inherently features nonlocality\cite{Ryu}. 
This nonlocality, which arises from quantum mechanical effects, is also expected to permute or spread information beyond the classical distortions of time and space mentioned earlier.
In fact, without this nonlocality, any CA-based discussion would likely have been rendered meaningless. 
For example, consider a metric that exhibits a time delay but no spatial distortion. 
In the absence of quantum effects, a CA encoding the information of the stretched horizon would be mapped identically onto the conformal boundary.
However, assuming there is a time delay, the time evolution on the conformal boundary should remain consistent with that of the stretched horizon while also appearing to reflect this delay. 
Yet, because the CAs themselves are entirely identical, its evolution would also remain the same, ultimately making the time delay disappear. 
Therefore, nonlocality is an essential property that must be addressed alongside the CA model.

Now, in the context of AdS/CFT, a possible interpretation of the commutation relation is that the $Z$ operator and the $T$ operator on the stretched horizon commute with each other.
Considering all the points so far, as shown in \figurename{1} a), it means that the results of time evolution at the stretched horizon and then decrypting to a conformal boundary are the same as those of time evolution after decrypting to a conformal boundary.
\begin{figure}
  \centering
  \includegraphics[width=\linewidth]{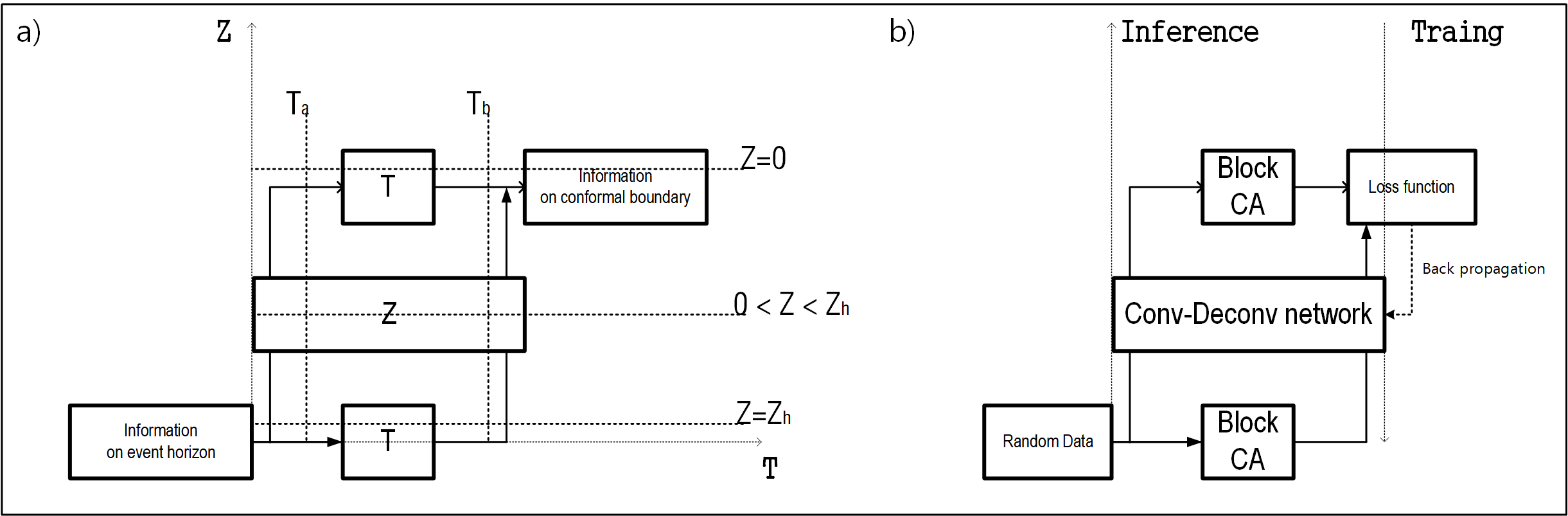}
  \caption{a)Commutativity in AdS space b)Deep learning algorithm}
  \label{FIG1}
\end{figure}
Moreover, the previously discussed requirement that the time delay and spatial distortion be reflected in the $Z$ function implies that the $Z$ function contains information about the curvature of the AdS spacetime.
This fact is represented in \figurename{1} a) also, where the bulk interval $0<Z<Z_h$ is shown to align with the location of the function $Z$.
This suggests that, just as general covariance served as a guide in formulating equations to compute curvature in classical theories, commutation relations could serve as a guide in constructing models for computing curvature with quantum effects in a theory of quantum gravity.

\section{A Deep Learning Algorithm for a Holographic System}
Let us now consider the problem of constructing a computational model that satisfies commutativity. 
In order to build such a model, as in the original work\cite{Hooft_1}, the amount of information defined on each hyperplane must remain finite. 
A straightforward and intuitive way to maintain finite entropy, while preserving zero curvature in Minkowski space, is to compactify the space into a torus.
Implementing this toroidal compactification on the black brane metric (cf.~Eq.~(2.5)) means imposing periodic boundary conditions, expressed via the following identifications:
\begin{equation}\label{eq:TorusIdentification}
\begin{aligned}
x &\;\sim\; x + L_x, \\
y &\;\sim\; y + L_y, \\
&\quad \text{(etc.\ for any additional coordinates).}
\end{aligned}
\end{equation}
thus effectively turning the black brane into a torus.
However, this compactification breaks the continuous Poincar\'e symmetry down to a discrete one. 
Note that Poincar\'e symmetry is not a necessary and sufficient condition for commutativity. 
As mentioned earlier, commutativity is an assumed condition from the perspective of information conservation, whereas Poincaré symmetry has been treated merely as a necessary condition for constructing the model in a straightforward manner.
If commutativity is truly a \emph{universal} property rather than a feature confined to a particular physical system, then it must continue to hold even if Poincar\'e symmetry is lost, i.e.\ even when the spacetime is compactified to a torus. 
Moreover, discussions of AdS/CFT and thermodynamics still remain valid, supported by the general arguments for topological AdS black holes\cite{Topo}.
Although the zero curvature still allows us to apply a CA model, the boundary conditions in the $T$ and $Z$ directions (for instance) introduce slight modifications to the paired evolution laws. 
Nonetheless, if we take the limit where the torus sizes go to infinity, Poincar\'e symmetry will be restored, enabling us to continue the previous line of reasoning without significant difficulty.

On the other hand, one might consider omitting compactification and instead restricting attention to a finite region of an infinitely large Minkowski spacetime, thereby ostensibly maintaining the symmetry locally while keeping the degrees of freedom finite. 
For instance, in the Ryu–Takayanagi prescription for the entanglement entropy of an AdS black hole\cite{Ryu}, one can choose a minimal surface that wraps the horizon partially, thereby relating the thermodynamic entropy at the horizon to the corresponding entanglement entropy in the CFT. 
Yet, because the entanglement entropy in a CFT is ultimately defined by the entanglement between a chosen region and its complement, it becomes challenging to quantify how local information at the horizon might be nonlocally distributed across the conformal boundary. 
Consequently, one must consider the entire conformal boundary, at which point global curvature arises and the situation becomes significantly more complicated.

Therefore, considering a setup that includes compactification, and assuming we know the time evolution as in a CA, let us implement a computational model to find the $Z$ operator using a deep learning algorithm.
In other words, assuming that $T$ is given, $Z$ will be trained as in \figurename{1} b).
The spatial dimension of the event horizon corresponds to the spatial dimension of the conformal boundary.
For visualization purposes, we represent the physical processes on two 3-dimensional spatial hypersurfaces encoded in the CA as 2-dimensional grids. 
Additionally, due to the correspondence, the entropy is the same, so we consider 2-dimensional grids of identical size.
For simplicity, classical CA will be considered. 
Although we employ a classical (block) CA as a toy model, one should note the fundamental differences between Boolean-state cellular automata and genuine quantum systems. 
In a quantum theory, states reside in a complex Hilbert space and evolve unitarily, thereby preserving norms and accommodating entanglement. 
By contrast, classical CA models typically involve discrete 0/1 updates without capturing complex amplitudes. 
While this paper draws upon CA-based intuition, fully realizing quantum features would require a more sophisticated, genuinely quantum cellular automaton framework.
However, note that nonlocality stemming from quantum effects manifests as the mixing and spreading of information, and it can be partially emulated even within classical CA models.
Note that our discussion of nonlocality concerns how quantum effects can spread or mix information in ways that exceed classical correlations. 
Meanwhile, in a classical CA, ‘nonlocal’ rearrangements often amount to permutations of discrete states. 
Although useful for illustrating certain aspects of holography, these permutations cannot fully replicate the physics of quantum entanglement. 
Thus, we caution readers that our classical CA remains an approximation, and genuine quantum nonlocality requires a more rigorous operator-based or entanglement-based formulation.

The computational model proposed in \figurename{1} b) corresponds to supervised learning in the domain of deep learning, but it can be observed that it exhibits a recursive structure where intermediate results during training serve as labels that act as ground truth in supervised learning.
To implement the proposed computational model, it is crucial to note that in deep learning algorithms, updating weights occur with each batch. 
Consequently, labels will be generated using the intermediate state of the model updated up to the given batch point. 
Therefore, as illustrated in the proposed algorithm below \figurename{2}, the loops should be divided into epoch loops and batch loops.
\begin{figure}[tb]
\centering
\caption{Training Loop}
\label{alg:training_loop}

\begin{algorithmic}[1]
  \State $X_{\text{train\_raw}} \gets 
    \text{GenRandBinTensor}(\text{trSize},\, wspan,\, hspan)$
  \State $X_{\text{train}} \gets 
    \text{PreprocData}(X_{\text{train\_raw}})$

  \State $epochs \gets \text{numEpochs}$
  \State $batches_{\text{per\_epoch}} \gets 
    \text{numBatchesEachEpoch}$
  \State $batch\_size \gets 
    \text{numSamplesPerBatch}$

  \For{$e \in \{0, \ldots, epochs-1\}$}
    \For{$b \in \{0, \ldots, batches_{\text{per\_epoch}}-1\}$}
      \State $T_{\text{raw}} \gets 
        \text{T}\bigl(\text{GetBatch}(X_{\text{train\_raw}},
        \;b,\;batch\_size)\bigr)$
      \State $T_{\text{proc}} \gets 
        \text{PreprocData}(T_{\text{raw}})$

      \State $Z_{\text{train}} \gets 
        \text{InterModel}\Bigl(\text{GetBatch}(
        X_{\text{train}},\;b,\;batch\_size)\Bigr)$

      \State $Z_{\text{pred}} \gets 
        \text{Logit2Pred}\Bigl(Z_{\text{train}},$
      \Statex \quad $\text{shape}=(-1,wspan,hspan)\Bigr)$

      \State $TZ_{\text{raw}} \gets \text{T}(Z_{\text{pred}})$
      \State $TZ_{\text{proc}} \gets 
        \text{PreprocData}(TZ_{\text{raw}})$
      \State $TZ_{\text{1hot}} \gets 
        \text{ConvertTo1Hot}(TZ_{\text{proc}},\,
        \text{nClasses})$

      \State $\text{TrainModel}\Bigl(
        T_{\text{proc}},\;TZ_{\text{1hot}},\;
        batch\_size\Bigr)$
    \EndFor
  \EndFor
\end{algorithmic}

\end{figure}

In the proposed algorithm, various functions are employed to streamline the training process. 
\texttt{IntermediateModel} refers to a model that has been trained up to the current batch, thereby allowing recursive learning when these intermediate states are placed inside a loop. 
The \texttt{GenerateRandomBinaryTensor} function produces raw random data to be used as model input. 
Meanwhile, \texttt{PreprocessData} ensures that this raw data is suitably normalized and transformed for the model. 
The \texttt{GetBatch} function extracts a specific batch from the dataset --- defined by the batch number and size --- by partitioning the data accordingly.
The \texttt{LogitToPrediction} function converts the model’s outputs (logits) into final predictions. 
This is typically achieved via softmax or sigmoid functions, which transform logits into probability values. 
\texttt{ConvertToOneHot} then encodes the label data into one-hot vectors, a common practice for multi-class classification. 
Finally, \texttt{TrainModel} takes both the input and target data to update the model’s parameters, 
minimizing the loss function over the course of training.

For the $T$ operator assumed above, at least the following two properties must be satisfied.
The first condition is locality, as in a typical CA model (to be distinguished from the nonlocality in AdS/CFT).
This is satisfied if the current cell is determined by the neighboring cells in the specified bound like Conway's game of life. 
Second, roughly speaking, to preserve quantum mechanical probability it should be time-reversible. 
That is, the cells in the past should be able to be determined from current cells.
Since two consecutive CA layers are needed to define the rule going to the future or the past, 
the evolution law for one time step can be split into a two-step procedure. 
Here, the block CA \cite{BCA} will be adopted as an example of time-evolution law. 
The well-known block CA rules are as follows.

\begin{figure}[tb]
\centering
\caption{Cellular Automaton Transition Rules for 2x2 Blocks}
\label{fig:cellular2x2}

\begin{algorithmic}[2]
\Require A 2x2 block of cells $(a, b, c, d)$ with states $0$ (dead) or $1$ (alive)
\Ensure Updated 2x2 block after applying transition rules
\State $S \gets a + b + c + d$ \Comment{Sum of live cells in the block}
\If{$S = 2$}
    \State \Return $(a, b, c, d)$ \Comment{Block remains unchanged}
\ElsIf{$S = 0 \text{ or } S = 1 \text{ or } S = 4$}
    \State \Return $(1-a, 1-b, 1-c, 1-d)$ \Comment{Flip all cells in the block}
\ElsIf{$S = 3$}
    \State $(a', b', c', d') \gets (1-a, 1-b, 1-c, 1-d)$ \Comment{Flip all cells in the block}
    \State \Return $(d', c', b', a')$ \Comment{Rotate the block by 180 degrees}
\EndIf
\end{algorithmic}

\end{figure}

For edge cells after the first rule is applied, it is assumed that cells on opposite sides are grouped together to form a torus to align with the compactification condition.
It is known that it satisfies locality by definition and is time-reversible.
Note that when we know the rules for even and odd sites like block CA, 
it is known that Hamiltonian can be expressed as an infinite sequence of commutators of these rules using Baker–Campbell–Hausdorff formula\cite{Hooft_2}.

On the other hand, we note that in order for the block CA rule to serve as a toy model of quantum time evolution, it should exhibit (at least approximate) linearity.
We can verify this linearity by showing that the block CA rule can be written in terms of a matrix multiplication.
In this toy model, the matrix is block diagonal, and although its elements can take binary values (0 or 1) at each discrete time step, we emphasize that each element is effectively time-dependent through iterative updates rather than remaining a fixed constant.
Of course, in a genuine quantum system, each matrix element would follow a continuous, complex-valued time evolution as governed by the Schrödinger equation, the Hamiltonian, and the initial condition, rather than collapsing to 0 or 1.
In this sense, the “linearity” of a classical CA is restricted to a local rule over \(\mathbb{Z}_2\), which remains fundamentally distinct from a unitary operator on 
a complex Hilbert space in a quantum system.

As a dataset, random input will be used in this paper.
A random dataset is defined on a stretched horizon, where the values 0's and 1's on each lattice might indicate the absence or presence of a photon at that position, respectively.
After mapping information to the lattice like this example, the $Z$ operator converts it to information in the corresponding conformal boundary.
Although discussions on how to encode physical states into information using bits as the unit are crucial in the proposed computational model, it is not the objective of this paper.

We now present the CNN architecture used to learn the operator $Z$ in \figurename{4}.
\begin{figure}
  \includegraphics[width=\linewidth]{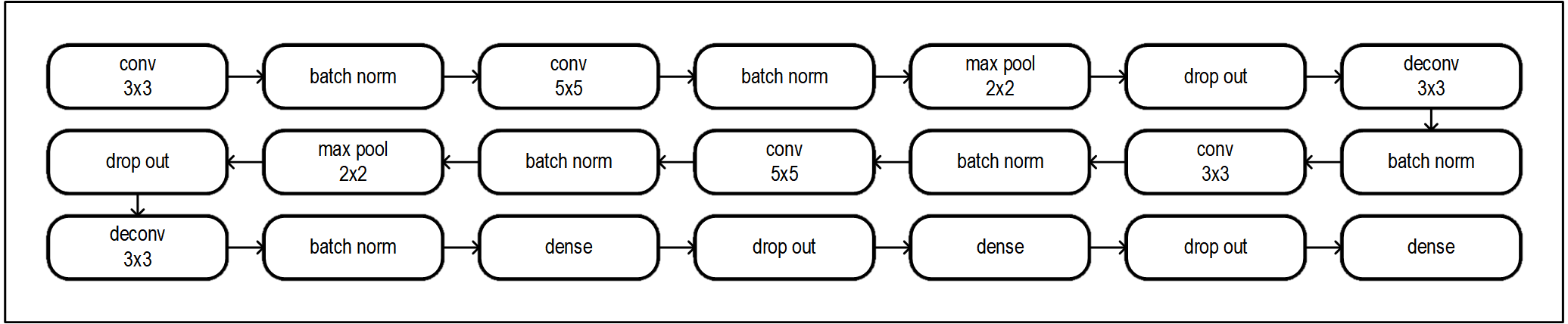}
  \caption{Convolution-deconvolution architecture}
  \label{FIG2}
\end{figure}
Here, a convolution-deconvolution\cite{CA} architecture was used recursively.
The proposed network follows a conventional deep learning architecture used in image segmentation, characterized by a convolution-max pooling-deconvolution block. 
The depth of these blocks is considered irrelevant to the current problem, as the correlation between the depth and the loss is not discernible critically. 
However, dropout layers and regularization are critically important for the problem we aim to solve. 
This is because, among the solutions that satisfy commutativity, there are cases where $Z$ is a constant function or trivial solutions.
Physically, these trivial solutions do not represent meaningful transformations in the AdS/CFT context. 
During the experimental process, it was observed that the model tended to learn trivial solutions in the absence of these techniques. 

However, more importantly, it is crucial to carefully select the loss function. 
As mentioned earlier, the proposed model tends to learn trivial solutions, so to avoid this, the loss function must be customized to penalize trivial solutions. 
In this model, we added a conditional statement to assign a very large loss value when a trivial solution is encountered, thereby encouraging the learning of non-trivial solutions. 
Nonetheless, non-trivial solutions are not necessarily physically meaningful. 
To obtain physically meaningful solutions, the model must incorporate the requirements derived from the holographic principle in the context of AdS/CFT.
For example, among the numerous solutions that $Z$ can take, the selected solution often depends on the weight initialization.
For more realistic learning, it is essential to initialize the weight based on a physically correct interpretation of the weight\cite{Koji_1}.
For instance, according to the interpretation of $Z$ presented earlier, the initialized weights could correspond to the classical solution of general relativity.
Then, the training can be interpreted as a correction due to quantum effects corresponding to nonlocality.
In a future, fully quantum framework, this approach could be extended by embedding unitarity or entanglement constraints directly into the architecture (e.g., through specialized layers or loss functions).

It can be confirmed in \figurename{5} that our architecture is able to learn the $Z$ operator.
\begin{figure}
  \centering
  \includegraphics[width=\linewidth]{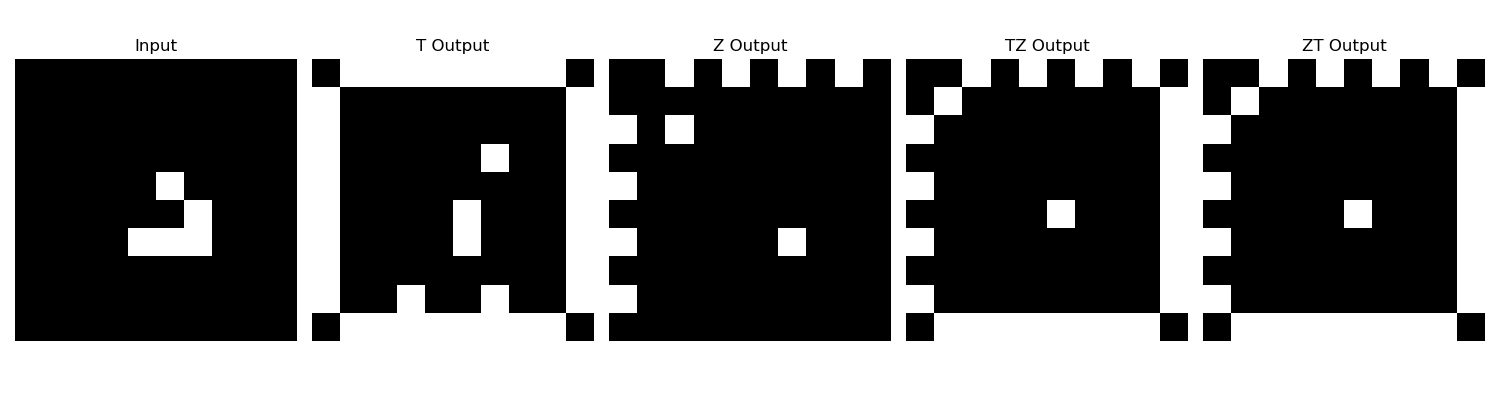}
  \caption{Training result}
  \label{FIG3}
\end{figure}
Here, white represents a living cell and black represents a dead cell.
During the inference process, the well-known glider pattern from the game of life was used as the input. 
However, it can be confirmed that the output after applying $TZ$ is the same as the output after applying $ZT$, even with different inputs.
Therefore, the proposed algorithm can learn a function that decrypts the information from one space to another by leveraging commutativity.
Because deep learning is a non-linear transform, this model could include non-trivial interactions if more physical constraints are added to the proposed model.

\section{Discussion}
In the proposed model, it was essential to assume that a time evolution law exists for a many-body system, as in cellular automata. 
Indeed, this rule is not as simple as block CA and can not be expressed in a mathematically closed form. 
Therefore, if this rule exists, how to find it is a difficult problem.
But again, one might also expect this rule to be discovered through a deep learning algorithm. 
First, we have a way to test a system in the standard model. 
Therefore, if one collects sufficient datasets via multiple experiments, the evolution law can be learned.
In fact, if we assume that random data really evolves according to the block CA rule as an example in the previous section, 
it has been confirmed that the block CA rule can be learned through an appropriate deep learning architecture and numerous random dataset.
Moreover, it can be confirmed that learning is possible in reverse time as well.

However, as previously noted, the curvature of the bulk causes the time evolution operator and the evolution operator along the holographic direction to form a closely related pair.
In fact, while the time evolution operators should be identical on the two hypersurfaces, the operator along the holographic direction must be capable of exhibiting a time delay.
Such a time evolution operator is unlikely to be a mere block CA. 
Nonetheless, identifying even a simple example of a paired set of evolution operators that satisfies these conditions could be a valuable step forward, offering a promising direction for advancing the current research.

Meanwhile, the proposed model does not represent the real world because it is made up of minimum simple assumptions.
In order to propose a model that is close to reality, more physical principles should be reflected in the model.
However, the advantage of deep learning algorithms is the fact that they are flexible in constructing the architecture.
For example, in order for the $Z$ operator to satisfy general covariance also, the weight sharing condition should be enforced in the proposed deep learning architecture\cite{Covariance}.
In addition, the activation function must be carefully selected to represent the CNN in covariant form.
For the case of G-CNN, activation function satisfies covariance when a pointwise function is used for activation.
In this way, it can be expected that mapping more physical requirements to the components of the deep learning architecture will bring the model closer to reality.

\end{document}